# Analyzing vehicle-pedestrian interactions: combining data cube structure and predictive collision risk estimation model

**Byeongjoon Noh[1] (powernoh@kaist.ac.kr), Hansaem Park[2] (saem@kaist.ac.kr) and Hwasoo Yeo[2]\* (hwasoo@kaist.ac.kr)**

[1]: Applied Science Research Institute, Korea Advanced Institute of Science and Technology, 291 Daehak-ro, Yuseung-gu, Daejeon, Republic of Korea

[2]: Department of Civil and Environmental Engineering, Korea Advanced Institute of Science and Technology, 291 Daehak-ro, Yuseung-gu, Daejeon, Republic of Korea

*corresponding author*

## Abstract

Road traffic accidents are a severe threat to human lives, particularly to vulnerable road users (VRUs) such as pedestrians causing premature deaths. Therefore, it is necessary to devise systems to prevent accidents in advance and respond proactively, using potential risky situations as one of the surrogate safety measurements. This study introduces a new concept of a pedestrian safety system that combines the field and the centralized processes. The system can warn of upcoming risks immediately in the field and improve the safety of risk-frequent areas by assessing the safety levels of roads without actual collisions. In particular, this study focuses on the latter by introducing a new analytical framework for a crosswalk safety assessment with various behaviors of vehicles/pedestrians and environmental features. We obtain these behavioral features from actual traffic video footages in the city with complete automatic processing. The proposed framework mainly analyzes these behaviors in multi-dimensional perspectives by constructing a data cube structure, which combines the Long Short-Term Memory (LSTM)-based predictive collision risk (PCR) estimation model and the on-line analytical processing (OLAP) operations. From the PCR estimation model, we categorize the severity of risks as four levels; "normal," "relatively safe," "warning," and "danger," and apply the proposed framework to assess the crosswalk safety with behavioral features. Our analytic experiments are based on two scenarios, and the various descriptive results are harvested; the movement patterns of vehicles and pedestrians by road environment and the relationships between risk levels and car speeds. Consequently, the proposed framework can support decision-makers (*e.g.,* urban planners, safety administrators) by providing valuable information to improve pedestrian safety for future accidents, and it can help us better understand cars' and pedestrians' proactive behavior near the crosswalks. In order to confirm the feasibility and applicability of the proposed framework, we implement and apply it to actual operating CCTVs in Osan City, Republic of Korea.

## 1. Introduction

The proliferation of advancements in information and communication technology (ICT) have led many cities worldwide to turn into smart cities by creating intelligent platforms and improving the quality of life in various fields such as environment, mobility, and safety. However, traffic accidents still arise numerous casualties; about 1.2 million fatalities and 50 million are injured every year [1], [2]. Road traffic accidents are a severe threat to human lives, particularly to vulnerable road users (VRUs) such as pedestrians, causing premature deaths [3]. They are exposed to various threats, such as drivers failing to yield when crossing [4]. In particular, risky situations frequently occur on crosswalks. The international bodies, such as British Transport and Road Research Laboratory and World Health Organization (WHO), pointed out that crossing a street, especially unsignalized crosswalks, is as dangerous as jaywalking [5]. Much research has been conducted on analyzing the severity and causal factors affecting vehicle-pedestrian collisions in order to stave off similar events after [6]–[9]. In general, they analyzed the actual data collected for a long-term period, and then came up with counterplans in order to prevent incidents such as the deployment of safety facilities in areas where events frequently occur. However, such approaches rely on historical data, which is even rare events, and are practically post-facto responses.

"Potential risk" has attracted intensive attention for the surrogate collision risk measurements to prevent actual collisions proactively, and analyzing this corresponds to supplement the current approaches [3], [10]–[15]. In general, a potential risk for collision between vehicle and pedestrian is a situation in which there is a certain probability of developing into an accident, such as non-yielding driving over a pedestrian, near-miss collisions, and overspeed in school zones [16]. There are two ways to prevent such collisions: (1) field process; and (2) centralized process. The former is an approach to respond immediately to risky situations in the field by sensing various information from the field equipment. Moreover, with the rapid growth of a cooperative-intelligent transportation system (C-ITS) in recent years, it has become possible to recognize upcoming risks and warn road users as a form of some field services; blind spot warning, abnormal situation recognition, etc. [17], [18]. For this, the authors in [19] proposed an intersection pedestrian collision warning system (IPCWS) that warns the drivers approaching the intersection by predicting the pedestrian's crossing intention with machine learning models. The authors in [20] predicted pedestrian-vehicle conflict 2 seconds ahead at signalized intersections. The long short-term memory (LSTM) neural network was used to predict these collisions, and this could be applied to collision warning system under C-ITS environment. These studies applied vision sensors such as the closed-circuit television (CCTVs), and make easier to obtain and handle the behaviors of vehicle/pedestrian on roads.

Meanwhile, the centralized process focuses on analyzing interactive situations between vehicle and pedestrian on road and obtaining valuable information about the risk factors threatening pedestrian safety. Moreover, it enables to assess the severity of risks in road levels, support decisions to improve the safety of the walking environment [21]–[24]. For example, the authors in [21] proposed a framework to evaluate pedestrian safety at unsignalized crosswalks using video data extracted through semi-automated processes. They analyzed pedestrian crossing behaviors, especially non-yielding maneuvers, based on vehicle trajectory speed and distance. The authors in [22] also used surveillance video footage, and demonstrated an effective way to identify risk levels, called key risk indicators that can offer predictive insights about pre-accident risk exposures, and evaluate risk severity. The authors in [23] investigated the relationships between pedestrians' crossing violation behaviors and the frequency of fatal pedestrian crashes in 55 signalized junctions. In addition, they considered the road design and signal settings to manage safety and convenience at the same time. Further, these approaches could develop into decision support systems (DSS) that can guide decision-makers such as urban planners and safety administrators to design and improve a walking-friendly environment by analyzing potential risks. The authors in [24] proposed road safety decision support system for analyzing road risks and possible countermeasures on a broad range in Europe by constructing a data cube model. They considered various features, largely divided into four categories; road users, infrastructure, vehicles, and post-impact care. This decision support system categorized the levels of risk factors. It can support to decide the order of priority of locations for deploying the safety facilities to improve the safety of road environment by analyzing the behaviors of road users, distributions of safety infrastructures, and responses of traffic accidents.

This study introduces a new concept of a pedestrian safety system that combines the field and the centralized processes. The proposed system warns of upcoming risks immediately in the field and helps to reinforce the safety of risk-frequent areas by assessing the safety levels of roads without actual collisions, as illustrated in **Figure 1**. The target of this concept is vehicle-pedestrian interaction nearby crosswalks. We studied the predictive collision risky (PCR) estimation model for vehicle-pedestrian interactions in our previous works as the field process [25]. This system can determine the severity of risk levels and warn the upcoming threats approximately 1-3 seconds ahead based on the video footage. This paper is an extension of our previous studies [25]–[28], aiming to describe the analytic methods of potential risky behaviors collected from multiple locations in the urban as the centralized process. Thus, descriptions of processing the video data and extracting behavioral features are omitted in this paper. Please refer to our previous studies for the detailed procedures [25]–[28].

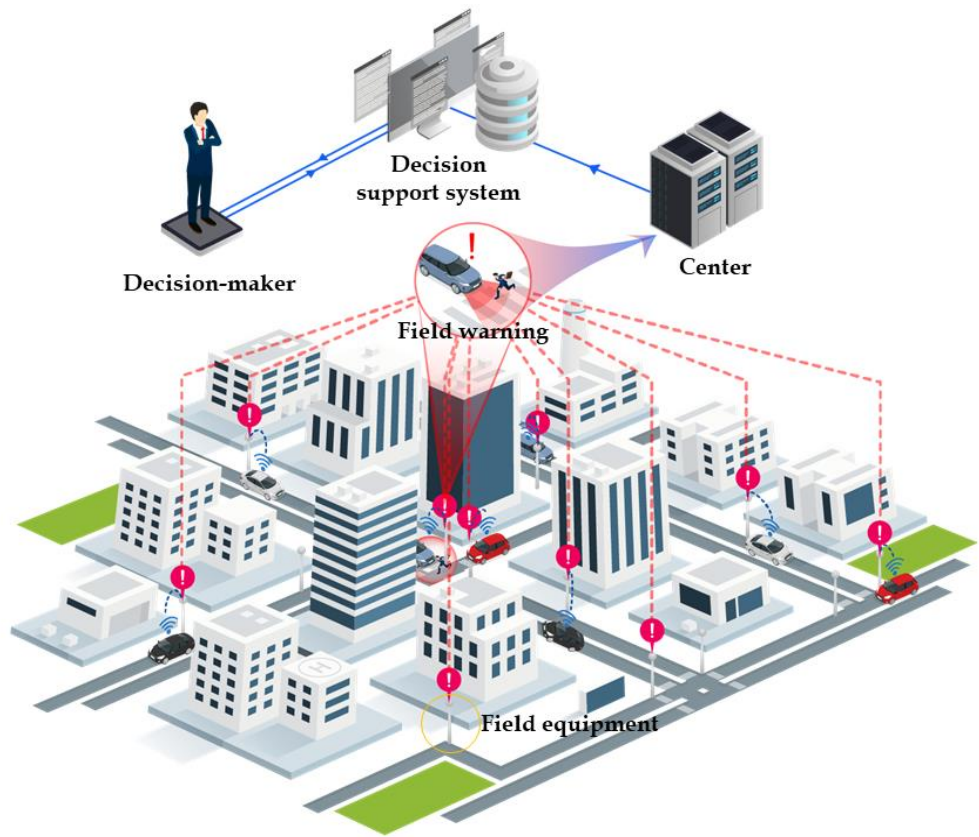

**Figure 1.** A new concept of pedestrian safety system that combines the field process and the centralized process

We propose a new analytical framework for a crosswalk safety assessment with various behaviors of traffic-related objects (*e.g.*, vehicle and pedestrian) and environmental features for the centralized process. The proposed framework mainly analyzes their interactions in multi-dimensional perspectives by constructing a data cube structure combining the LSTM-based PCR estimation model and on-line analytical processing (OLAP) operations. These behaviors are extracted from real traffic video footage from the CCTVs in the city, which are accumulated in a data warehouse over an extended period. These behaviors are automatically extracted by using computer vision techniques [26], [27], [29] without any human observations manually, which require inefficiently prolonged time and high labor costs. From the LSTM-based PCR estimation model, we categorize the severity of potential risks as four levels; "normal," "relatively safe," "warning," and "danger." These levels are used to assess crosswalk safety, and also to analyze the behavioral features. The data cube model and OLAP operations can analyze these behaviors comprehensively by considering various features together, such as types of situations, road environments, changes in vehicle speeds, and pedestrian positions [30]. For example, the results of analyses could give answers or clues for the following questions: "How do cars behave at the risk level 3?" and "why did more risky situations occur in the school zone even if the vehicles move slowly?". Furthermore, with the LSTM-based PCR estimation model, the proposed framework enables finding the meaningful information without actual collisions and providing an immediate field alert.

The main objectives of this study are: (1) to extract objects' behavioral features, such as speeds of vehicles and pedestrians, pedestrian safety margin, and PCR levels, from video in multiple locations during long-term periods; (2) to design a data cube, called *SafetyCube2*, by combining with the PCR estimation model; and (3) to comprehensively analyze the scenes (situations) nearby crosswalks in multiple perspectives using OLAP operations with various abstraction levels. This framework allows seeing which factors affect the risks, especially collision risks of vehicle-pedestrian interactions. Consequently, the proposed framework can support decision-makers, such as urban planners and safety administrators, by providing valuable information to improve pedestrian safety for future accidents and better understand how cars and pedestrians behave near the crosswalks proactively. To the best of our knowledge, this is the first report on designing a data cube structure combining with a deep learning-based predictive risk measuring model and handling a large amount of vision-based behaviors of vehicles and pedestrians in multi-dimensional perspectives. We confirm the feasibility and applicability of the proposed framework by applying it to the actual operating CCTVs on roads in Osan City, Republic of Korea.

The remainder of this paper consists of three chapters described as follows:

1. Materials and methods: Descriptions of data sources, PCR estimation method, multi-dimensional data cube model, and OLAP operations for vehicle-pedestrian interaction analysis.

2. Experiments and results: Case studies based on two scenarios by analyzing vehicle-pedestrian interactions according to locations, severity of risk levels, and other factors, and discussion of results and limitations of the research.

3. Conclusions: Summary of our study and future research directions.

## 2. An analytic framework for road safety assessment using vehicle-pedestrian interactions

### 2.1. Overall architecture of the proposed framework

This section introduces an overall proposed framework for road safety assessment analyzing the interactive behaviors between vehicles and pedestrians. This framework enables handling many objects' behavioral features from multiple crosswalks using the diversified OLAP operations at various abstraction levels. **Figure 2** illustrates the overall architecture of the proposed analytic framework. This framework mainly consists of three parts: (1) data sources and preprocessing; (2) feature extraction; and (3) multi-dimensional analysis. In the first part, we obtain the trajectories of vehicles and pedestrians by scenes (situations) from the CCTV cameras over the roads. In order to detect and segment traffic-related objects, we use a mask regional-convolutional neural network (mask R-CNN) [31], which is widely used for object detection in computer vision. Then, the detected objects are traced by using the deep simple object real-time tracking (deep SORT) algorithm, one of the object trackers [32], [33]. Finally, since typical CCTV cameras record the video in oblique view, the obtained objects' trajectories are transformed into top view.

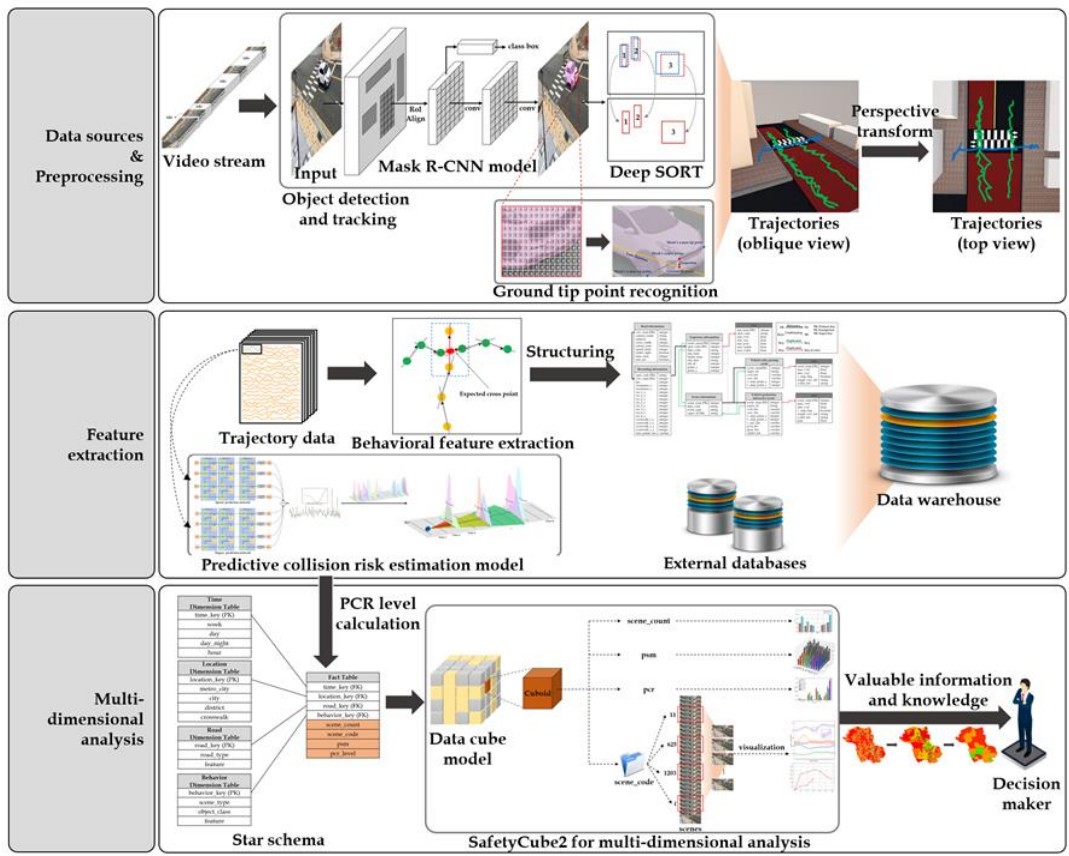

**Figure 2.** The overall architecture of the proposed analytical framework

We extract various behavioral features affecting the collision risks in the feature extraction part, such as car speed, pedestrian position, and the pedestrian safety margin (PSM). Furthermore, we estimate

PCR by using the deep LSTM-based model in this part. The PCR information is used to warn of upcoming risky situations immediately on the road site as field process (see **Figure 1)**. The feature extraction part aims to structure a data warehouse with behavioral features and external databases such as road information. As a core methodology of this study, the multi-dimensional analysis aims to construct a data cube model and analyze vehicle-pedestrian interactions using OLAP operations. This analytic method can elicit valuable information for decision-makers to reinforce the safety of the road environment without great expense or new additional infrastructure specifically to extract travelers' behaviors.

## 2.2. Data sources and preprocessing

Our experiment uses the datasets based on the video footage from CCTV cameras provided by Osan Smart City Integrated Operations Center in Osan City, Republic of Korea. These cameras are originally used to surveil and deter street crime instances, but we apply them to analyze vehicle-pedestrian interactions. We selected cameras deployed over crosswalks in various environments such as the school zone and unsignalized crosswalks, and then collected the videos from nine crosswalks for 14 weekdays from January 9th to January 28th during rush hours; morning peak time (from 8 am to 10 am) and night peak time (from 6 pm and 8 pm). These spots are located near residential complexes and schools, and they have a high floating population. Due to the privacy issue in the Republic of Korea, we processed the video footage by deploying a computing server locally in the Osan center and retained and viewed the processed trajectories after removing the origin video. **Figure 3** and **Table 1** describe the views of our test spots and characters of these locations and recording meta-data, respectively.

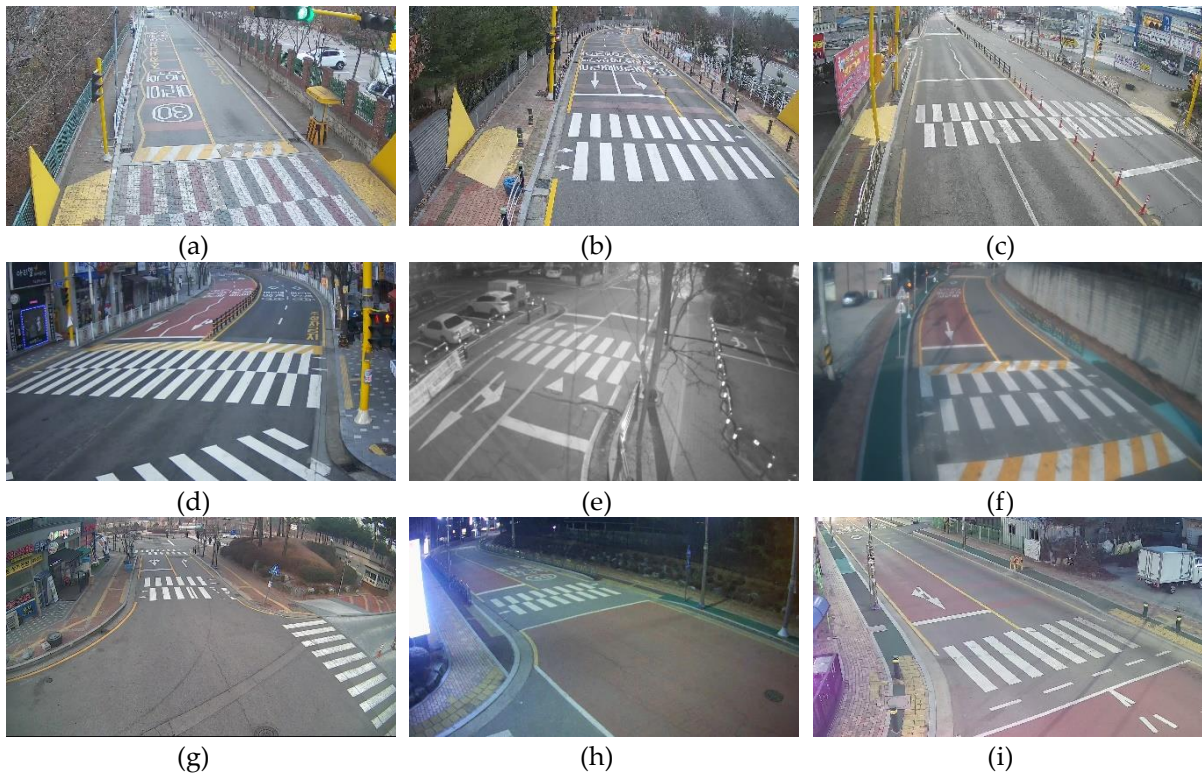

**Figure 3.** Camera views of our test spots (a) Spot A to (i) Spot I, respectively

In the preprocessing part, three steps are conducted: (1) object detection and tracking; (2) ground tip point recognition; and (3) perspective transform. First, we detect and trace the traffic-related objects using the mask R-CNN model and deep SORT algorithm. In parallel, ground tip points of each object are recognized in oblique views. Since typical CCTV cameras record oblique views, it is necessary to find the contact points of objects in top view in order to extract behavioral features such as speeds and positions reliably. Thus, they are transformed into points in top views, called contact points, in the perspective transform step. The contact points are basepoints when extracting objects' behavioral features, so ground tip points of vehicles and pedestrians are points on the ground directly underneath the front bumper and on the ground between the feet, respectively. Please refer to our previous works [26], [27], for detailed preprocessing procedures. With these sets of points, we can obtain the trajectories and classify the scenes into two types: "single-object scene," where only one type of object passes, and "interactive scene," where both types of objects (vehicle and pedestrian in our case) are involved at the same time.

**Table 1.** Information of our test spots

| Spot code | Spot name and camera number | Crosswalk length (*m*) | School zone | Speed cam. | # of lanes | Signal light | Speed limit (*km/h*) |
|-----------|----------------------------|------------------------|-------------|------------|------------|--------------|----------------------|
| A | Unam Elementary school, back gate #2 | about 8 | √ | x | 2 | √ | |
| B | Yangsan Elementary school, main gate #1 | about 11 | √ | x | 3 | √ | |
| C | Gohyeon Elementary school, back gate #2 | about 20 | √ | x | 4 | √ | |
| D | Daeho Elementary school opposite side #3 | about 23 | √ | √ | 4 | √ | |
| E | Municipal Southern Welfare/Daycare center #3 | about 7 | √ | x | 2 | x | 30 |
| F | iFun daycare center #2 | about 8 | √ | x | 2 | x | |
| G | Segyo complex #9 back gate #2 | about 8 | x | x | 2 | x | |
| H | iNoritor daycare center #2 | about 8 | √ | x | 2 | x | |
| I | Kids-mom daycare center #3 | about 7 | √ | x | 2 | x | |

**Note**: √: Yes, x: No.

As a result of data collecting and preprocessing, such as detecting objects, recognizing their trajectories, and extracting their behavioral features, we obtained 45,890 scenes with movements of the traffic-related objects, as seen in **Table 2**.

**Table 2.** The numbers of the obtained scenes after preprocessing

| Spot code | The number of scenes (after preprocessing) | | | |
|---|---|---|---|---|
| | Collection time | Car-only passing scene | Vehicle-pedestrian interactive scene | Total number |
| A | Day | 1,875 | 1,133 | 4,221 |
| | Night | 806 | 407 | |
| B | Day | 1,253 | 786 | 2,908 |
| | Night | 468 | 401 | |
| C | Day | 1,658 | 1,125 | 4,111 |
| | Night | 663 | 665 | |
| D | Day | 4,525 | 852 | 6,955 |
| | Night | 1,969 | 241 | |
| E | Day | 2,657 | 1,714 | 3,876 |
| | Night | 1,976 | 608 | |
| F | Day | 1,564 | 883 | 7,587 |
| | Night | 917 | 512 | |
| G | Day | 2,541 | 1,714 | 5,612 |
| | Night | 992 | 365 | |
| H | Day | 1,457 | 875 | 2,845 |
| | Night | 386 | 127 | |
| I | Day | 2,853 | 2,660 | 7,775 |
| | Night | 1,719 | 543 | |

## 2.3. Feature extraction

In this section, we describe which features (variables) are used to analyze vehicle-pedestrian interactions. The used features are categorized into two types: (1) external features; and (2) behavioral features. The external features include the external factors that are likely to affect potential risks, such as road types, school zone designations, and safety structures (fences, bumper, red urethane, etc.). We define the behavioral features as factors likely to affect potential risks posed by the vehicle and pedestrian movements such as speed, acceleration, position, stop before the crosswalk, PSM, and PCR level. Finally, these features are categorized again into individual features and interactive features according to their attributes for configuring star schema, which is required for constructing a data cube model.

In addition, the use of PSM and PCR level as surrogate measurements enables to assess potential risks, not actual collisions. PSM is one of the surrogate measurements assessing interactions between vehicles and pedestrians [34]–[36]. In general, the definition of PSM is a time difference from when a pedestrian passes a certain point (as virtual conflict point) and when the next vehicle reaches the same point [37], [38]. Thus, PSM can be calculated as $T_2 - T_1$ where $T_1$ is a time when pedestrian reaches a certain point and $T_2$ is a time when a vehicle arrives at the same point. If a car has passed before the pedestrian, PSM is a negative value, but if the PSM value is close to zero regardless of its sign, it means there is less margin to avoid a collision at the conflict point. Unlike PCM, which is calculated based on the entire trajectories already acquired, the PCR level can proactively warn against collisions and prevent accidents. the PCR level only uses previous trajectories and current positions of vehicles and pedestrians to yield severity of risks. The behavioral features used to design the data cube model in this study are described in **Table 3**, and behavioral features are extracted automatically from the video footage using computer vision and deep learning techniques. The additional description for the PCR level is in section 2.3.1, and others are explained in our previous studies [28], [39] in detail.

**Table 3.** The extracted behavioral features used to design data cube model

| Category | Target object | Feature name | Format | Description | Example |
|---|---|---|---|---|---|
| Individual behavior | Vehicle | speed | list | • Vehicle speeds by time <br> • Unit: *km/h* | • [14.3, 12.0, 9.8, …] |
| | | average speed | numeric | • Average vehicle speed in one scene <br> • Unit: *km/h* | • 14.5 |
| | | position | list | • Vehicle positions based on crosswalk by time <br> • Represented as: "before crosswalk", "on crosswalk", or "after crosswalk" | • [before crosswalk, on crosswalk, after crosswalk] |
| | | acceleration | list | • Vehicle accelerations by time <br> • Represented as: "acceleration", "deceleration", or "no change" | • [acceleration, no change] acceleration, deceleration, no change] |
| | | crosswalk distance | list | • Distances between vehicle and crosswalk by time <br> • Unit: *m* | • [4.1, 3.3, 1.9, …] |
| | | stop behavior | str | • Whether the vehicles stopped before passing the crosswalk <br> • Represented as: "stop" or "no stop" | • stop <br> • no stop |
| | Pedestrian | speed | list | • Pedestrian speeds by time <br> • Unit: *km/h* | • [2.3, 2.9, 1.2, …] |
| | | average speed | numeric | • Average pedestrian speed in one scene <br> • Unit: *km/h* | • 3.1 |
| | | position | list | • Pedestrian positions by time <br> • Represented by: "sidewalk", "crosswalk", or "CIA (crosswalk influenced area)" | • [sidewalk, CIA, sidewalk] <br> • [sidewalk, crosswalk, sidewalk] |
| | | crosswalk distance | list | • Distance from crosswalk by time <br> • Unit: *m* | • [3.1, 2.0, 1.3, …] |
| Interactive behavior | - | relative position | list | • Relative positions between vehicle and pedestrian by time <br> • Represented as: "front" or "behind" <br> • "Front" means pedestrian is in front side of the car <br> • "Behind" means pedestrian is rear side of the car | • [behind] <br> • [behind, front] |
| | | vehicle-pedestrian distance | list | • Distances between vehicle and pedestrian by time <br> • Unit: *m* | • [4.1, 3.3, 1.9, …] |
| PSM | numeric | ● | Pedestrian safety margin | ● | 3.2 |
| | | ● | Unit: *sec.* | ● | -1.5 |
| PCR level | numeric or category | ● | Severity of levels for potential collision risk | ● | danger |
| | | ● | Based on the predictive collision risk | ● | level 2 |
| | | ● | Represented as: "danger" (level 4), "warning" (level 3), "relatively safe" (level 2), or "normal" (level 1) | | |

* CIA (crosswalk influenced area): CIA refers to the road area adjacent to the crosswalk, where pedestrians often enter while crossing the road. Some people cross the street on the correct-zebra-marked zone as well as around of this area.

### 2.3.1. Measurement of predictive collision risk level between vehicle and pedestrian

As one of the behavioral features of vehicles and pedestrians, we use PCR levels as a surrogate measurement assessing the severity of potential risks between them. In order to obtain PCR levels, we conduct two steps: (1) trajectory prediction using the deep LSTM network; and (2) predictive collision risk area estimation. The LSTM network, an evolution of recurrent neural network (RNN), to cope with the time-based sequential data such as trajectories. In this model, we predict the trajectories (positions of objects) after few seconds by using their speeds and degrees. Then, we estimate PCR areas by using confidence intervals of speeds and degrees in order to compensate that the predicted positions have some tolerances, not exactly where they fit. With the lower and upper bounds (confidence interval) for speed and degree, the areas are in the form of a predictive collision risky area (PCRA). Our experiment predicts the objects' trajectories after about 1, 2, and 3 seconds. Consequently, we can obtain the multiple PCRAs after about 1, 2, and 3 seconds (see **Figure 4 (a)**). Then, we defined the PCR level as a measurement of severity of risk level by using the degree of overlap of PCRAs between vehicle and pedestrian as follows:

● Danger: When PCRAs for after 1 second for vehicle and pedestrian are overlapped

● Warning: When PCRAs for after 2 seconds for vehicle and pedestrian are overlapped

● Relatively safe: When PCRAs for after 2 seconds for vehicle and pedestrian are overlapped

● Normal: Not overlapped

When determining the severity of risk levels, danger case always takes precedence over other levels (see **Figure 4 (b)**)

The amount of time it takes a driver to react to an upcoming threat is called perception-intellection-emotion-volition (PIEV) time [40]. Perception is time to discern an object or event, intellection is time to understand the implications of the object's presence or event, emotion is time to decide how to react, and volition is time to initiate the action, such as engaging the brakes. Much research has been conducted on measuring the safe PIEV time, and this has a range depending on a variety of conditions such as the road conditions, visibility, driver's characteristics [41]–[45]. The America Association State Highway and Transportation Officials (AASHTO) recommended value of PRT equal to 2 till 2.5 seconds [40], and the authors in [41], [46] concluded as 1.48 to 2.5 seconds. Thus, we categorized the severity of risk levels by about 1 second unit, and decision-makers can adjust this depending on the environmental conditions.

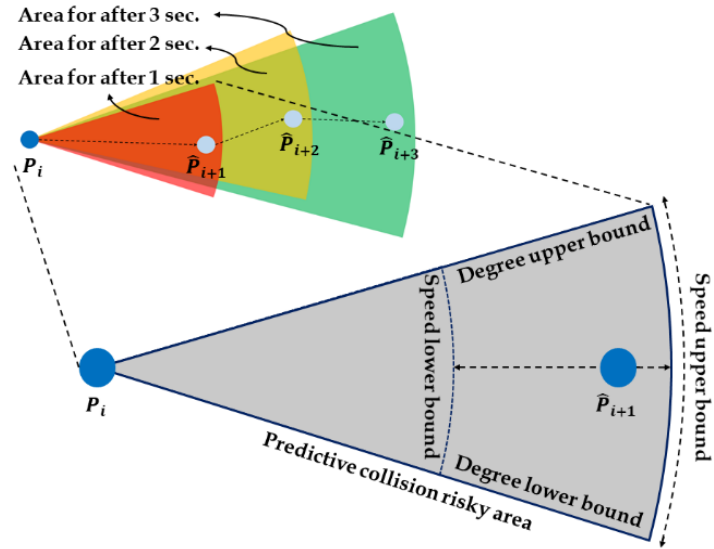

(a)

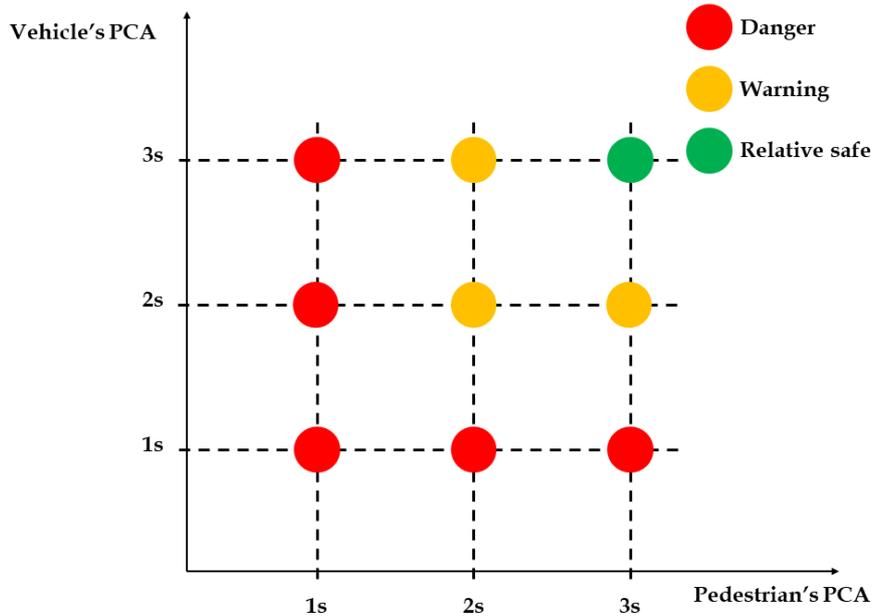

(b)

**Figure 4.** Definition of (a) PCRAs for after about 1, 2, and 3 seconds; and (b) PCR levels

## 2.4. Multi-dimensional analysis

### 2.4.1 Data cube model for vehicle-pedestrian interaction in multiple perspectives

This section handles analytic methodologies for vehicle's and pedestrian's interactive behaviors using external features in multi-dimensional perspectives. We extracted a large amount of data accumulated in multiple areas during long-term periods on an urban scale in previous parts. In order to handle, model, and view these bulks of data in a variety of dimensions, a data warehouse requires on more concise, flexible, and subject-oriented schema than traditional relational databases (RDB) [47], [48]. The key is to construct a data cube model with diversified multi-dimensional schemas. In general, it is a schema defined by dimension tables, and a fact table called a star schema resembling a starburst with the dimension tables shown in an outspread pattern around the central fact table. Dimension table is the entity's perspective for which a user wishes to maintain records, a bulk of data, and can be easily added if a new analysis viewpoint is required. The fact table is a large central table connected by each dimension table via key values [49] and contains measurements used to quantify the fact values.

In this study, the star schema is designed for analyzing vehicle-pedestrian interactions in multiple perspectives, called *SafetyCube2*. Similar to *SafetyCube* version 1 [39], this cube also has four dimensions (location, time, road, and behavior), but different four measurements: (1) scene count; (2) PSM; (3) PCR level, and (4) scene code; as seen in **Figure 5**. "Scene count" is based on an aggregation of the number of scenes and means the summation (or ratio) given by each dimension. "PSM" and the "PCR level" are also aggregated by PSM values and PCR levels obtained in the feature extraction part, respectively. PCR levels are represented the numeric values for aggregation, as 1 (normal), 2 (relatively safe), 3 (warning), 4 (danger). For example, if there are two scenes with warning and danger levels in a particular spot, the PCR level of this spot can be calculated as 2.5 arithmetically. Unlike others, "scene code" is not directly aggregated, as illustrated in **Figure 6** but used to view and comprehend the changes of objects' behaviors by time (frame) in scenes that satisfy the conditions resulted from OLAP operations. Further, it enables scrutinizing the vehicle-pedestrian interactions and providing a better understanding of how objects behaved near the crosswalks.

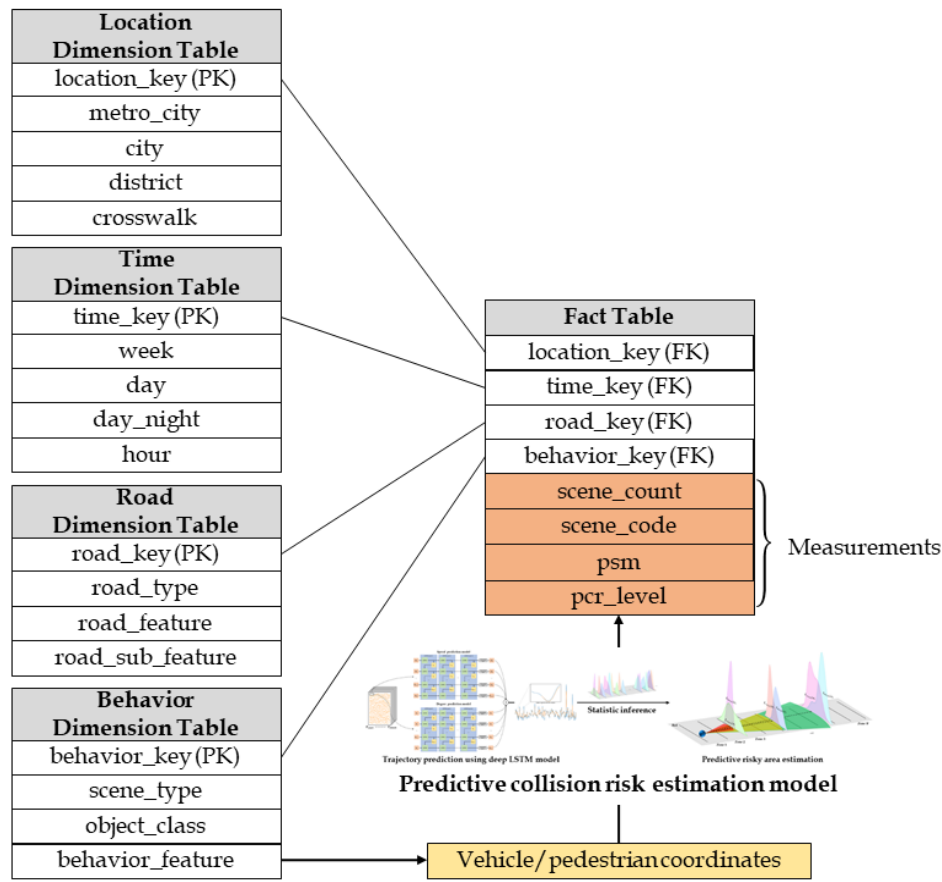

**Figure 5.** Star scheme to construct data cube model

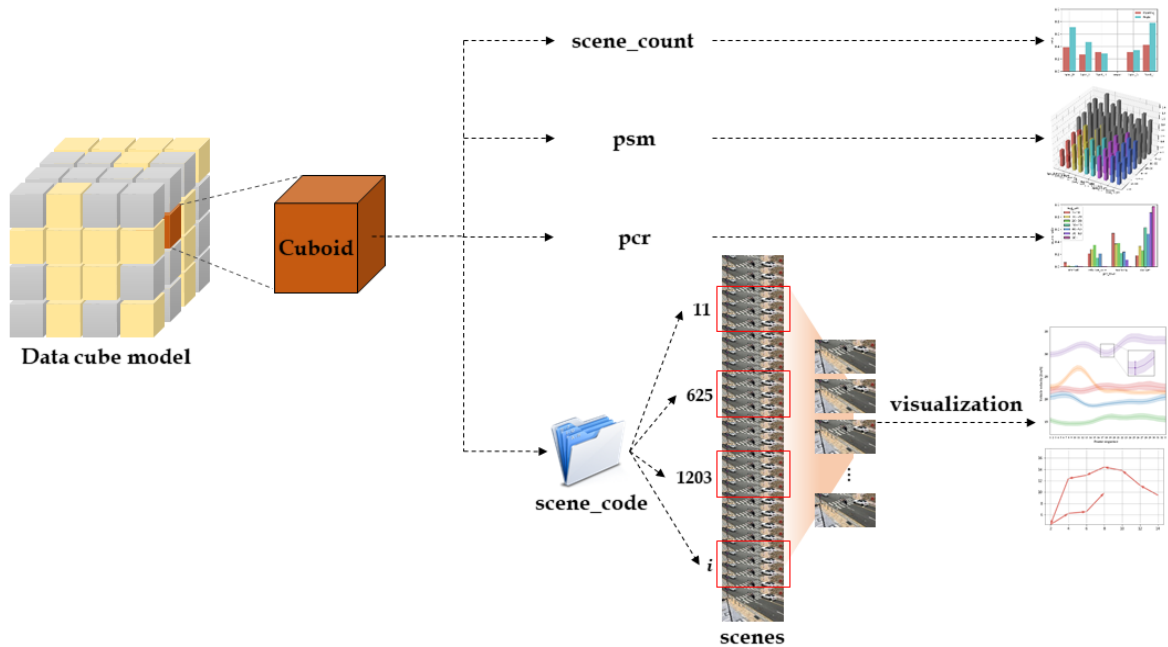

**Figure 6.** Ways to analyze vehicle-pedestrian interactions using each measurement in data cube model

### 2.4.2 Concept hierarchies and OLAP operations

One of the core functions of the data cube model is that it organizes multiple dimensions, and makes it possible to analyze the given data in multiple abstraction levels. Its concept hierarchy is in the form of a hierarchical tree structure and allows data to be managed at multiple abstraction levels by mapping a set of low-level concepts to high-level and more generalized concepts [47]. **Figure 7** shows four concept hierarchies defined in this study: (1) location dimension, (2) time dimension, (3) road dimension, and (4) behavior dimension. **Figure 7 (a)** shows hierarchy of location dimension ordered by levels of administrative districts, "spot < neighborhood < district < city < Metropolitan city (province) < all". For example, "Spot C" is the lowest level, "spot," and "Gyeonggi-do" is the highest level, "province," of location dimension respectively. **Figure 7 (b)** and **Figure 8** show the hierarchy of administrative districts and the map of Osan City, which is our test spot, respectively. The time dimension is ordered by "hour < day_night < day < week < all" as seen in **Figure 7 (c)**. Similarly, road and behavior dimensions are segmented as "road sub-feature < road feature < segment < all" and "behavioral feature < object type < situation sub-type < situation type < all." In the road dimension (see **Figure 7 (d)**), we define that "road segment" is road types such as intersection and crosswalk, and "road feature" is largely separated as signalized and unsignalized roads. "Road sub-feature" is the lowest level of road characteristics such as whether each road is designated as a school zone or not and whether speed camera, fence, and red urethane are deployed. Finally, **Figure 7 (e)** shows the hierarchy of behavior dimensions, categorized by scene types (refer to preprocessing part), and the lowest level contains the behavioral features described in **Table 3**.

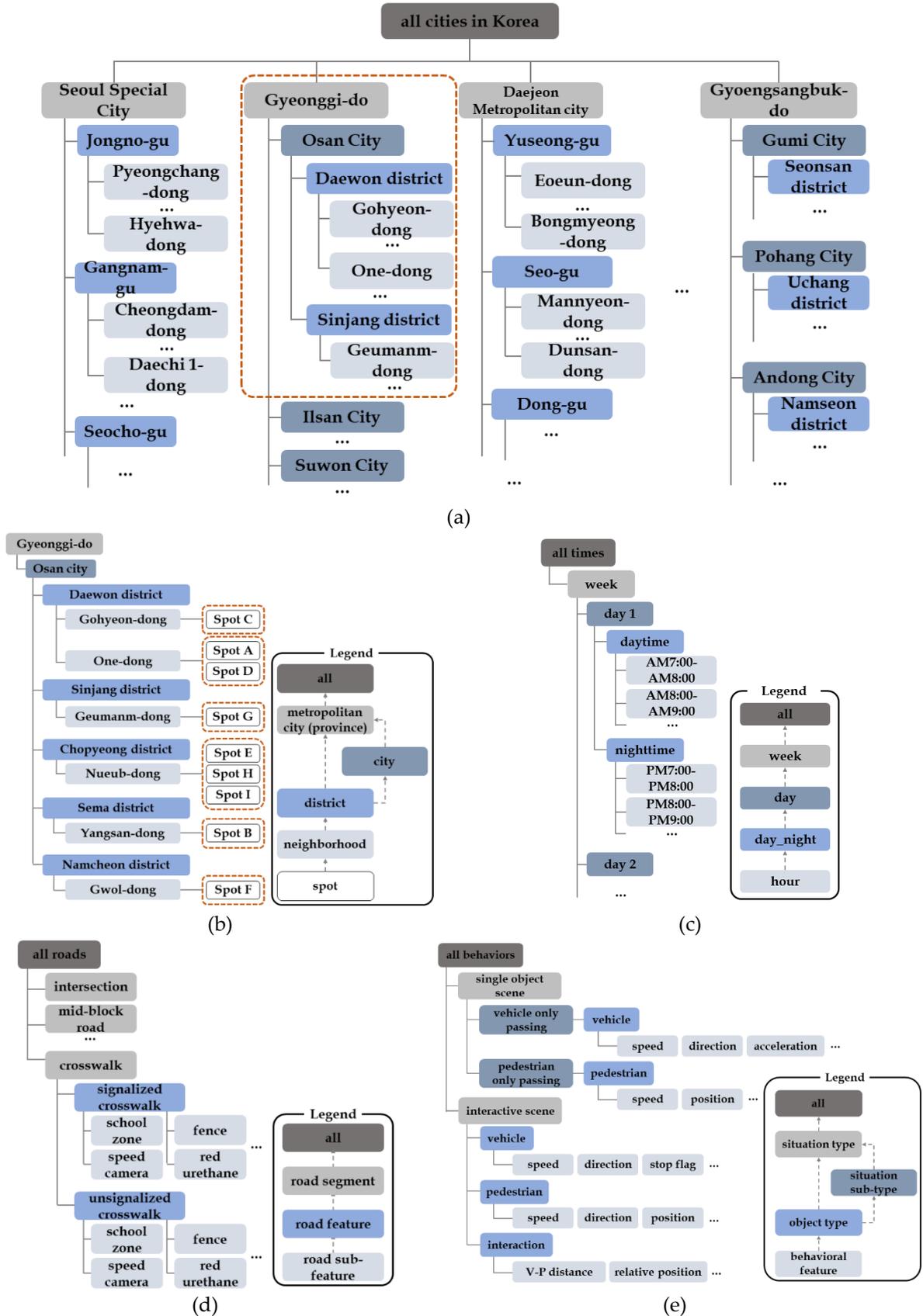

**Figure 7.** Concept hierarchies of data cube: (a) location dimension (for all cities in the Republic of Korea); (b) location dimension (for Osan City in Gyeonggi-do); (c) time dimension; (d) road dimension; and (e) behavior dimension

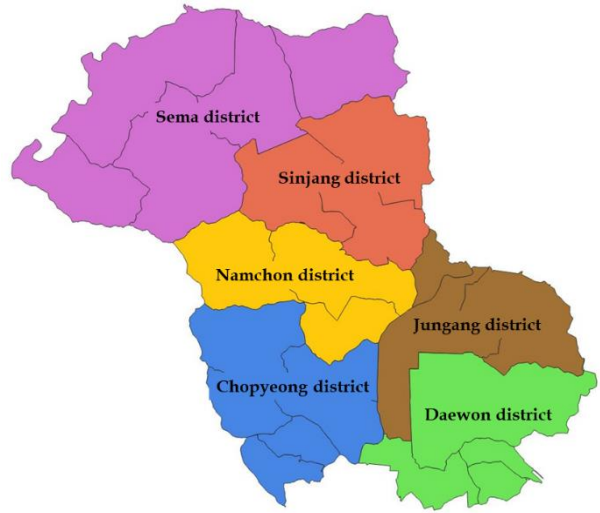

**Figure 8.** Osan City map

Through these multiple concept hierarchies, the proposed data cube model allows interactive querying and investigating the available data with flexibility from various perspectives [30]. For flexible, extensible, and interactive data analysis, OLAP operations are adopted. These provide a user-friendly environment for analysis [50]–[53] and support retrieving useful information. There are mainly five OLAP operations: (1) Roll-up; (2) Drill-down; (3) Dice; (4) Slice; and (5) Pivot. Roll-up operation generalizes the aggregated data cube by climbing up a concept hierarchy for a dimension, whereas drill-down is the reverse of roll-up. This operation specifies the aggregated data cube from less detailed data to more detailed data. This operation can be realized by either stepping down a concept hierarchy for a dimension or introducing additional dimensions. Dice operation is used to select two or more dimensions in the form of a sub-cube, and slice operation is used to select data by fixing one dimension of the given cube. Finally, the pivot operation rotates the axis of data to provide a substitute presentation of data. In our experiments, MySQL database management system (DBMS) is used to store the data in the data warehouse, and Python is used to extract features, including the PCR level, and scrutinize vehicle-pedestrian interaction by constructing data cube model for querying, analyzing, visualizing, and reporting the results. Then, the users can analyze the data to facilitate multi-dimensional views and create reports, depending on the requirement, by selecting the desired dimensions.

## 3. Experiments and Results

### 3.1. Experimental design

This section describes how to analyze the vehicle-pedestrian interactions based on behavioral and external features in the multi-dimensional perspectives using the data cube model. Before the main analyses, we briefly investigate the extracted behavioral features; average vehicle speed, PSM, and car stopping behavior. **Figure 9** shows the average vehicle speed by scene types and locations. Spots A, B, C, and D are signalized crosswalks, and others are unsignalized crosswalks. This figure shows that vehicles tend to travel faster when no pedestrians are present (car-only passing scenes) and slower when there are pedestrians (interactive scenes). In addition, the average speeds of vehicles in Spots C and I exceed the speed limit (30 km/h). As seen in **Figure 9**, since Spot C has four lanes, drivers seem to become insensible awareness of the speed limit. However, the average speeds of Spot F vehicles, which have similar road characteristics to Spot C, are lower than those in Spot C. This seems to be related to the deployment of a speed camera. Spot I also has an average overspeed, but there are no safety facilities such as speed cameras, signal lights, and fences. Thus, Spot I can be categorized as a dangerous location in this simple analysis, and we can hypothesize that a speed camera can suppress overspeed.

Next, **Figure 10** is a boxplot for PSM in each unsignalized crosswalk. Since there is no signal phase information, it is challenging to investigate subtle vehicle-pedestrian behaviors in signalized crosswalks. Thus, we focus on analyzing their interactions in unsignalized crosswalks for PSM and car stopping behavior. It should be noted that a negative PSM value means that a vehicle passed before pedestrian, and smaller the absolute value means there are fewer margins to avoid a collision. This figure shows that the ratios of non-yielding situations (negative values) and the yielding situations (positive values) are similar in all spots. This result does not figure out where the dangerous area is, but we conclude that it is better to design experiments by separating its sign when analyzing PSM. The main analysis for PSM is conducted in Section 3.3.

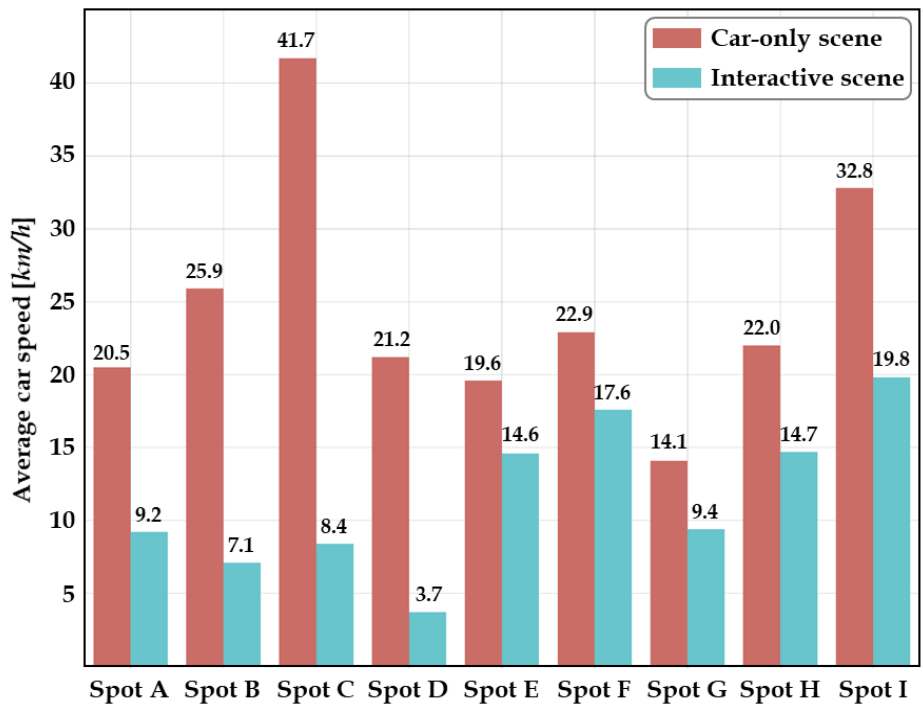

**Figure 9.** Average car speed by scene types in each spot

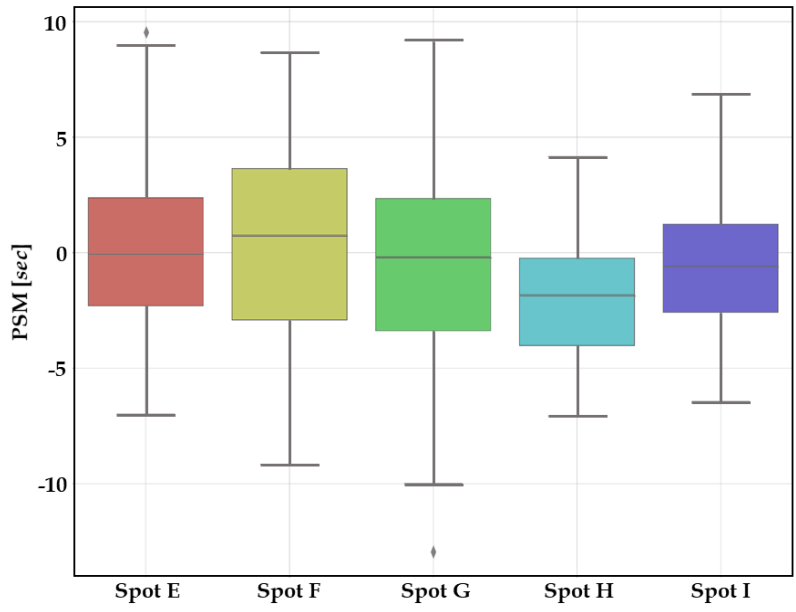

**Figure 10.** Distributions of PSMs in each unsignalized crosswalk

Finally, to investigate how many cars stop before passing the crosswalk when pedestrians are on the streets, we plotted the percentages of "stop" and "no stop" behaviors in each unsignalized spot as illustrated in **Figure 11**. In all unsignalized spots, more than half the drivers did not stop before passing the crosswalks. However, Spot H has the highest stopping percentages than others. It is believed that this area has various safety facilities such as a fence, red urethane, and the "school zone" mark. Meanwhile, Spot F fails to stop when pedestrians were on the crosswalk, despite the designation of the school zone. Thus, this spot seems to need proactive responses to encourage stopping for pedestrian priority.

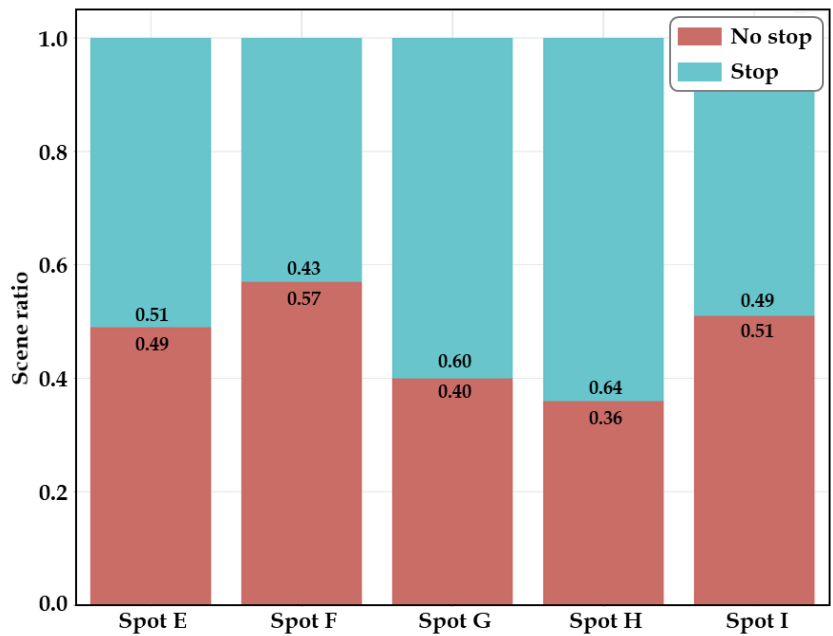

**Figure 11.** Percentages of "stop" and "no stop" behaviors in each unsignalized spots

Our main experiments are based on data cube and OLAP operations and scrutinize their interactions in multi-dimensional perspectives. A system manager attempts to adjust and choose dimensions depending on the analytic purposes and preferences in various viewpoints and levels of abstractions. Furthermore, we validate the feasibility and applicability of the proposed analytic framework by performing experiments and describing their results. The adopted analysis scenarios are as follows:

- Scenario I: Analysis of vehicle-pedestrian interactive scenes where vehicles are not yielding in unsignalized crosswalks using PSMs

- Scenario II: PCR level-based analysis for vehicle-pedestrian interactive scenes in and outside the school zones

## 3.2. Scenario I: Analysis of vehicle-pedestrian interactive scenes where vehicles are not yielding in unsignalized crosswalks using PSMs

In scenario I, we analyze vehicle-pedestrian interactive scenes to understand situations based on PSMs and other features. Note that PSMs at signalized crosswalks greatly depend on the traffic signal at the time of encounter, so this scenario aims to analyze in unsignalized crosswalks; Spots E, F, G, H, and I. First, we investigate the ratios for yielding and non-yielding situations in each spot. We can distinguish these situations by using signs of PSM values; positive values mean vehicles are yielding, negative values mean non-yielding. Further, in order to establish the relationships between PSMs and other features, we choose time and road dimensions; time period, and fences deployment as safety facilities, respectively. The OLAP operations to derive results that satisfy conditions are as follows:

**Drill-down on** Time (from "*all*" to "*day_night*")
**Drill-down on** Location (from "*all*" to "*spot*")
**Drill-down on** Road (from "*all*" to "*road_sub_feature*")
**Drill-down on** Behavior (from "*all*" to "*scene_type*")
**Dice for** (**Measure** = "*scene_count*") and (**Time** = ["*daytime*" | "*nighttime*"] in day_night) and (**Location** = "*all spots*" in Osan City) and (**Road** = "*fence*" in unsignalized crosswalk) and (**Behavior** = "*interactive*")
**Slice on** Scene ("*psm*" < 0)

**Figure 12** shows percentages of non-yielding scenes (negative PSMs) by time period in each spot. Since negative PSM values imply that the vehicle fails to yield to a pedestrian in the crosswalk, this generally presents more risk than positive values. As seen in this figure, Spots E, F, and H have fences, others have not. We can observe that there are higher proportions of non-yielding scenes at nighttime than daytime in most spots except for Spot H. In particular, Spot I has the highest non-yielding ratios in nighttime even if there is no fence. This area looks dangerous, and additional safety facilities need to be deployed proactively.

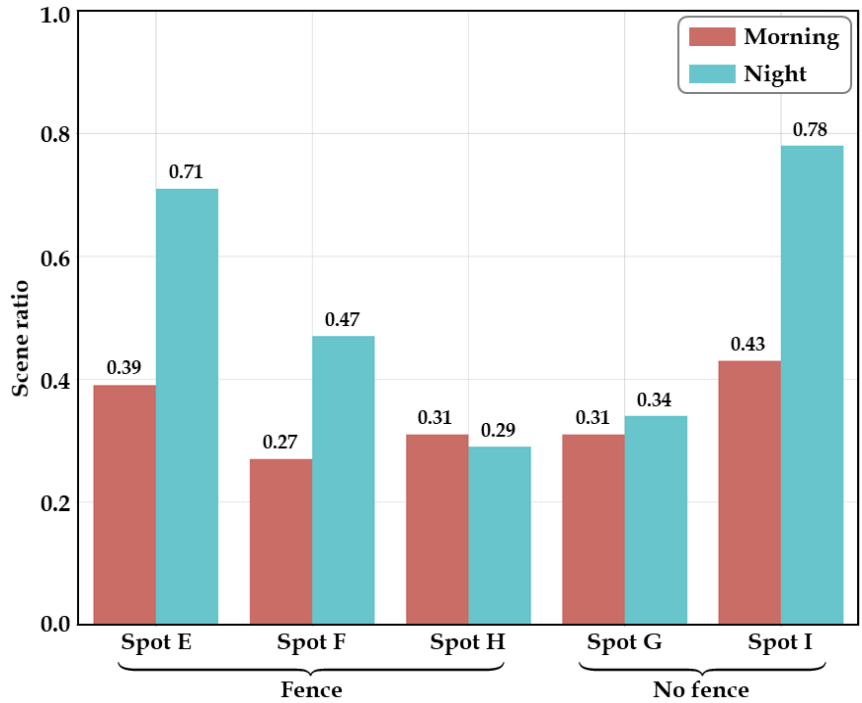

**Figure 12.** Percentages of non-yielding scenes (negative PSMs) by time period in each spot

To analyze collision risks in each area more in-depth, we spread distributions of PSM values for non-yielding cases in each spot. In fact, collision risk increases as PSM approaches zero regardless of PSM values' sign. For this, the OLAP operations are conducted from **Figure 9** as follows:

**Dice for** (**Measure** = "*psm*") and (**Time** = ["*daytime*" | "*nighttime*"] in day_night) and (**Location** = "*all spots*" in Osan City) and (**Road** = "*fence*" in unsignaled crosswalk) and (**Behavior** = "*interactive*")
**Slice on** Scene ("*psm*" < 0)

**Figure 13** shows distributions of negative PSM values in each spot and time. PSM values are close to zero in night peak time, making it look more dangerous than daytime. The reason behind is that the data collected from around 6 pm to 8 pm has low visibility as it represents data after the sunset in the Republic of Korea. Meanwhile, in relationships of non-yielding and the PSM value, Spots E, F, and I have high non-yielding ratios in nighttime (see **Figure 12**), whereas Spots E and F have sufficient PSM values about 1 second, while Spot I has minimal (closed to 0) PSM values. To sum up the two analyses, Spot I has many potential risky situations and needs to take action proactively.

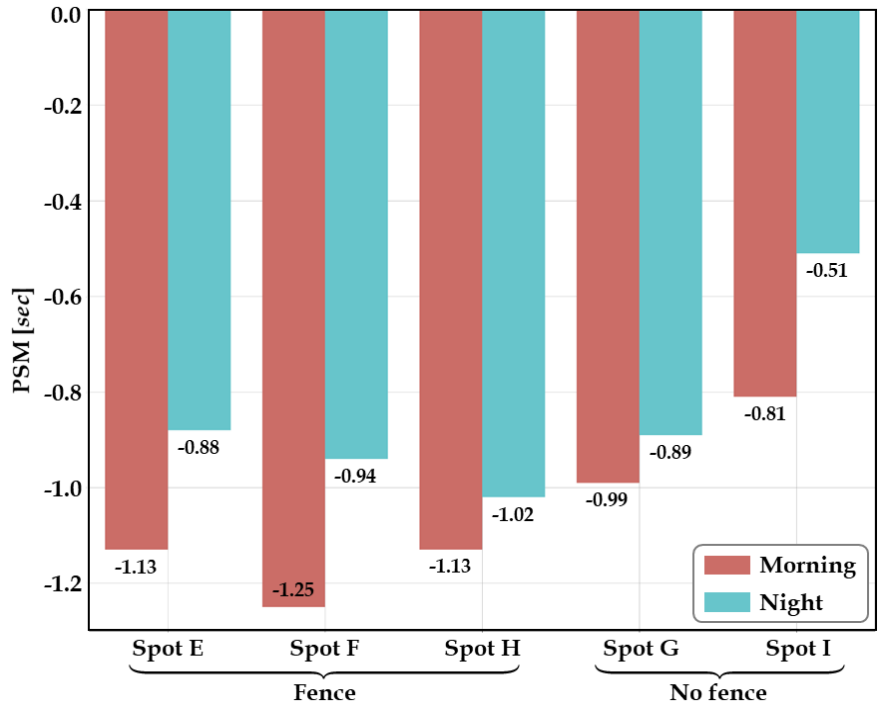

**Figure 13.** Distributions of negative PSMs by time period in each spot

At this point, we target to analyze risky behaviors in Spot I. The speed of the vehicle is one of the most influential factors in PSM value, so we plot PSM values by average vehicle speed separated by 10 *km/h*, as follows:

**Dice for** (**Measure** = "*psm*") and (**Time** = ["*daytime*" | "*nighttime*"] in day_night) and (**Location** = "*all spots*" in Osan City) and (**Road** = "*fence*" in unsignalized crosswalk) and (**Behavior** = "*average car speed*" in car feature in interactive scene)

**Slice on** Location (spot = "Spot I")

**Slice on** Scene ("*psm*" < 0)

In comparison with other spots' results, we also arrange the plot satisfying the same conditions in Spot F. **Figure 14 (a)** and **(b)** represent the PSM plots in Spot I and F, respectively. As described in **Figure 14 (a)**, we can observe two facts: 1) it is more dangerous at night peak time than during day peak time at the same car speed; and 2) the faster the speed, the closer the PSM value is to zero. On the other hand, Spot F also has PSM values closer to zero as the average car speed increases (see **Figure 14 (b)**), but the absolute values tend to be greater than the values of Spot I. Since Spot I does not have any safety facilities deployed to prevent collisions, unlike Spot F, it seems more dangerous.

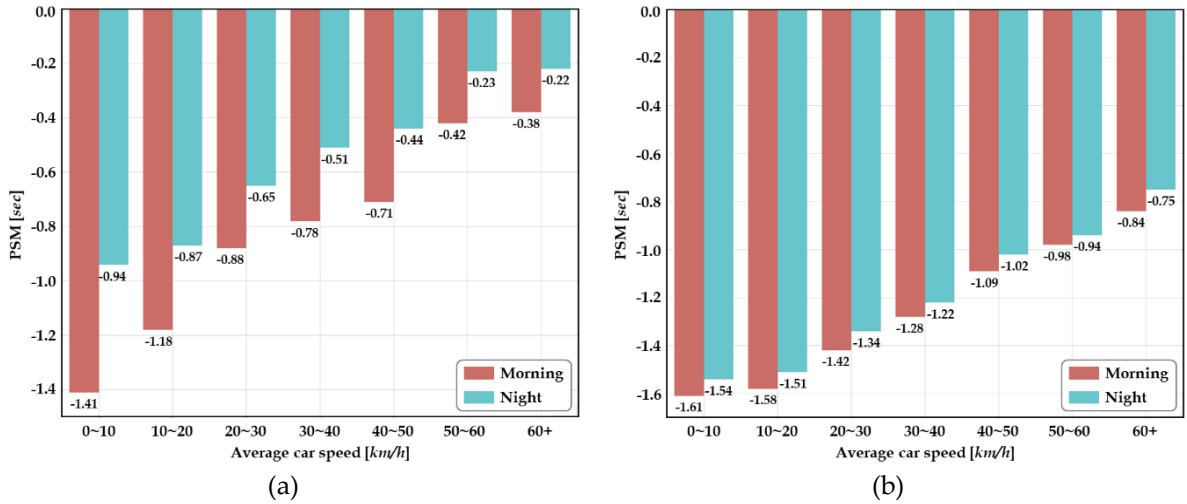

**Figure 14.** Distributions of negative PSMs by average car speeds in (a) Spot I and (b) Spot F by time period

To sum up, in this scenario, we analyze vehicle-pedestrian interaction at unsignalized crosswalks with various features such as time period and safety facilities. First, we focus on investigating if the drivers yielded to the pedestrians crossing the street in each spot at daytime and nighttime. As a result, there are more situations at nighttime where drivers do not yield to the pedestrians passing first than those at daytime. Next, we examine PSMs for non-yielding cases (negative values) to understand margins for collisions. Spot I has minimal absolute values in both daytime and nighttime. In detail, we distribute PSM values by average vehicle speeds separated into 10 $km/h$ and compare two spots; Spot F (the highest absolute PSM on average) and Spot I. Spot F has some safety facilities, but Spot I have none of them. We cannot guarantee that the behaviors in Spot F are safer due to safety facilities, but we can obtain useful information that the deployment of safety facilities is essential in Spot I.

In this study, we automatically extract PSM, as one of the surrogate measurements for potential risks such as near-miss collisions, by using computer vision techniques. In this scenario, we perform PSM-based analysis to investigate vehicle-pedestrian interactions, such as a basis for determining the concession behaviors between them. Note that the main objectives of this study are to measure the severity of potential risk levels in advance, warn of upcoming threats to drivers and pedestrians, and use them, collected from urban scale to analyze risks comprehensively. However, it is difficult for the PSM value to obtain such information because it is just a time difference between the vehicle and pedestrians passing through a certain point. Thus, the following section focuses on analyzing interactions based on the PCR level, considering their previous trajectories and predictive risks.

### 3.3. Scenario II: PCR level-based analysis for vehicle-pedestrian interactive scenes in and outside school zones

This scenario focuses on analyzing how vehicle-pedestrian interactive behaviors differ between the school zone and non-school zone. In the Republic of Korea, typical school zones are designated as an area where there are facilities for under age 13, including elementary schools, tutoring academies, daycare centers, and the drivers who cause the accident in these areas are severely punished. For example, hurting or killing children in these areas can be fined up to 3,000 million won or imprisoned for life [54]. In addition, typical school zones have road safety structures to ensure safe movements for children to prevent traffic accidents.

In our experiment, the target spots for the school and non-school zones are Spot E and Spot G, respectively. They are chosen as target spots because they have similar road environments; unsignalized crosswalks, same number of lanes, no deployment of speed cameras, and same speed limits. In addition, there are similar numbers of interactive scenes, about 2,322 and 2,079 in Spots E and G, respectively (see **Figure 15**).

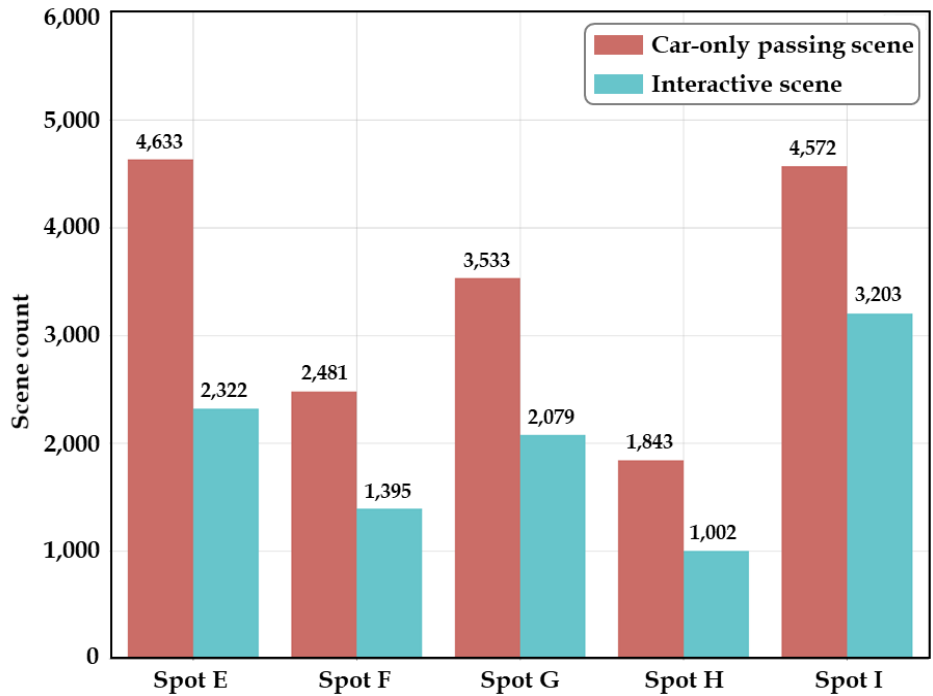

**Figure 15.** Scene counts by scene type in each spot

First, we compare scene ratios in these spots by average car speed as following OLAP operations:

**Drill-down on** Location (from "*all*" to "*spot*")
**Drill-down on** Road (from "*all*" to "*road_feature*")
**Drill-down on** Behavior (from "*all*" to "*behavioral_feature*")
**Dice for** (**Measure** = "*scene_count*") and (**Time** ="*all*") and (**Location** = "*all spots*" in Osan City) and (**Road** = "*unsignalized*" in crosswalk) and (**Behavior** = "*avg. car speed*" in interactive scene)
**Slice on** Location (spot = ["*Spot E*" | "*Spot G*"])

In **Figure 16**, red bars represent the results of Spot E (school zone), and others represent the results of Spot G (non-school zone). In this figure, we can observe that most cars travel slowly under the speed limit regardless of the school zone designation, and even there are more ratios of scenes with 0~10 *km/h* of vehicles in the non-school zone than those in the school zone. From the road character perspective, the car speeds are higher than in the school zone because of the deployment of fences. It is equipped in the school zone to separate the roads from the sidewalks, but drivers may not mind pedestrians entering the road suddenly.

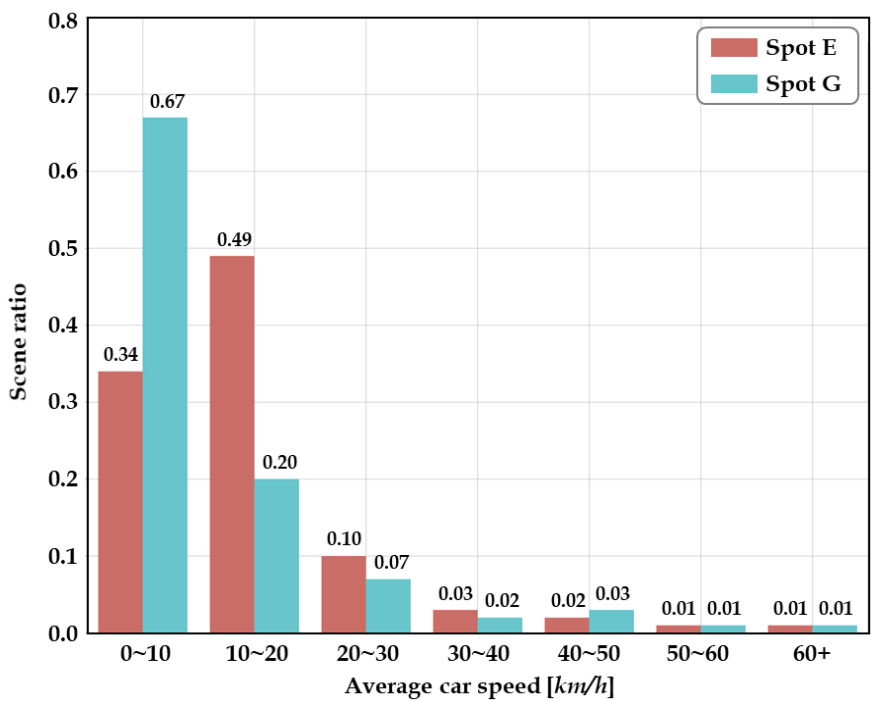

**Figure 16.** Scene ratios by average car speed in school zone (Spot E) and non-school zone (Spot G)

However, it is difficult to determine if these areas are actually dangerous by using average car speed alone. At this point, we can plot a graph for the scene ratios by the PCR level in both spots as follows:

**Dice for** (**Measure** = "*scene_count*") and (**Time** ="*all*") and (**Location** = "*all spots*" in Osan City) and (**Road** = "*unsignalized*" in crosswalk) and (**Behavior** = "*pcr level*" in interactive scene)
**Slice on** Location (spot = ["*Spot E*" | "*Spot G*"])

**Figure 17** shows the scene ratios in Spots G and E by PCR levels. In Spot E, the school zone, most situations are "normal" or "relatively safe." In our experiment, we define a "normal" situation as a situation in which the areas expected to be positioned after three seconds of a car and a pedestrian do not overlap at all, and a "relatively safe" situation is a situation in which these areas are overlap within two to three seconds. On the other hand, Spot E has a relatively higher proportion of warning and danger situations. This result does not mean that the area will be safer because it is designated as the school zone. However, given this result, it can be seen that travelers passing through the school zone in a similar environment are less exposed to dangerous situations, and proactive actions would be needed in Spot G to prevent a vehicle-pedestrian collision.

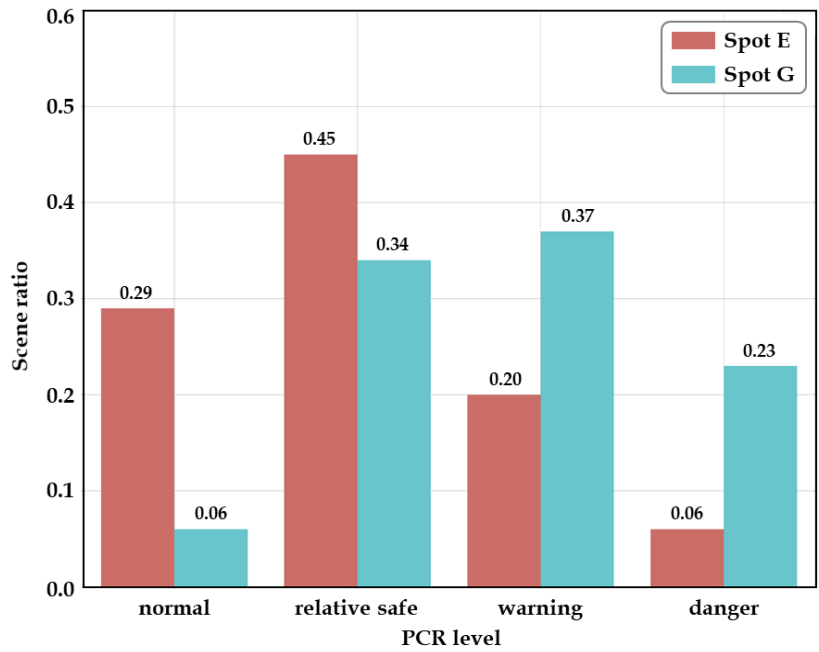

**Figure 17.** Scene ratios by PCR levels in Spots G and E

For more in-depth analysis for the PCR level, we investigate the average car speed as follows:

**Dice for** (**Measure** = "*scene_count*") and (**Time** ="*all*") and (**Location** = "*all spots*" in Osan City) and (**Road** = "*unsignalized*" in crosswalk) and **Behavior** = ("*pcr level*" and "*avg. car speed*") in an interactive scene)
**Slice on** Location (spot = ["*Spot E*" | "*Spot G*"])

**Figure 18 (a)** and **(b)** illustrate percentages of scenes by average car speed at each PCR level in each spot. Average car speed is also separated by 10 *km/h*. We anticipate that the more dangerous the situation, the greater the proportion of the average car speed. However, the results in the school zone and non-school zone are completely different.

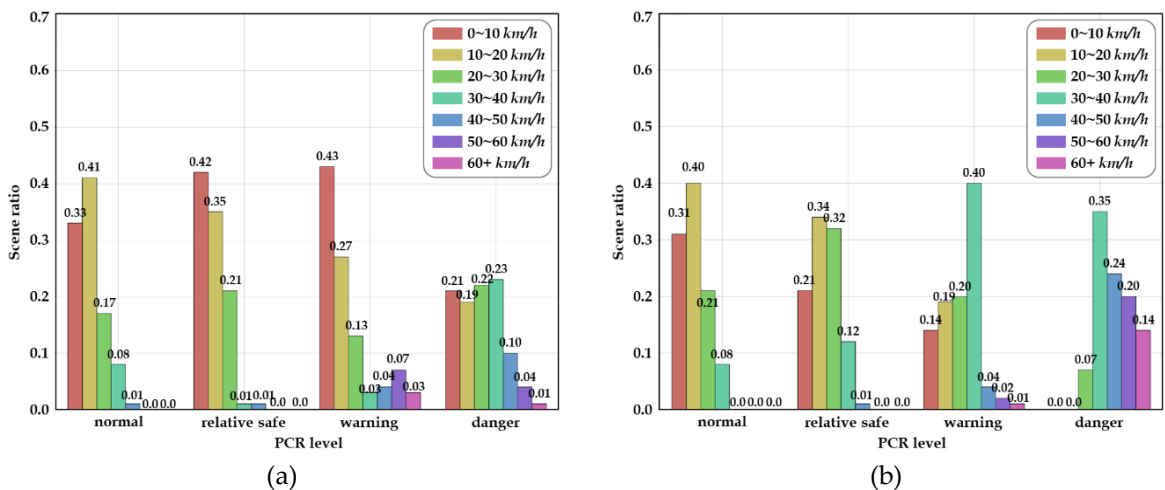

(a)                                                                (b)

**Figure 18.** Scene ratios by average car speed at each PCR level in (a) Spot G and (b) Spot E

**Figure 18 (b)** shows that there are higher proportions of scenes with low speeds of vehicles in "normal" and "relatively safe" levels, whereas there are higher proportions of a scene with high speeds of vehicles in "warning" and "danger" levels as. Meanwhile, **Figure 16** and **Figure 17** represent that most cars move slowly under the speed limit and the total percentage of "warning" or "danger" levels in Spot E is approximately 25%. These results show that this crosswalk seems to be relatively safe. However, unlike these results, **Figure 18 (a)** shows that major proportions of risky situations at "warning" and "danger" levels are scenes with low car speed. Most pedestrians in this area (school zone) would be children, and they may have suddenly jumped into the road without paying attention.

In summary, we compare vehicle-pedestrian interactions in Spot E (school zone) and Spot G (non-school zone) in this scenario. As a result, most vehicles travel slowly under the speed limits in both spots, which seem safe. Next, we investigate the PCR level in both spots. The school zone has high percentages of "normal" and "relatively safe" situations, and the non-school zone has mainly "warning" and "danger" situations. So far, Spot E and Spot G look safe and dangerous, respectively. However, the proportions of scenes according to average car speed in each PCR level show different interpretable results. In the school zone, the "warning" level has the high proportions of scenes with low speeds, unlike the case of the non-school zone. Therefore, it can be interpreted that the children in the school zone may have suddenly jumped into the road, so proactive action is needed to prevent these sudden appearances.

### 3.4. Discussions

The three contributions of this research are: (1) to extract objects' behavioral features from the video footage in multiple locations during long-term periods and organize hierarchical data structure; (2) to organize a data cube model combining the deep LSTM-based PCR estimation model; and (3) to comprehensively analyze the vehicle-pedestrian interactions in multiple perspectives using OLAP operations with various levels of abstraction in order to understand their behaviors and prevent accidents proactively. Much research has been conducted on handling potential collision risks of vehicles and pedestrians and analyzing their interactions. In **Table 4**, we compare the previous literature for analyzing road accidents with our approach. The authors in [55], [56] analyzed the actual traffic accident by using history and expert views, and explored main factors affecting accidents. Unlike such conventional approaches, the authors in [57]–[59] and the proposed framework applied vision-based data to analyze behavioral interactions. One of the closest approaches to ours is [60], which similarly extracted objects' behaviors from their trajectories and analyzed interactions at unsignalized roads, especially secondary interactions, which are important factors in understanding crossing behaviors.

Our process conducts a multi-dimensional analysis to investigate objects' subtle behaviors by various conditions along with road characteristics, and time period. Further advanced from our previous version, *SafetyCube* version 1 [39], the proposed analytic framework also combines the PCR estimation model based on the deep LSTM network, so it can consider collision risks with predictive positional areas of vehicles and pedestrians in few seconds ahead. Moreover, the proposed framework applies the PCR levels to measure the severity of situations and assess road safety without actual collisions along with other features. Consequently, the proposed framework can provide policy decision-makers with ability to improve the safety of the road environment for future collisions, especially for pedestrians by scrutinizing their interactions and then gaining a better understanding of how they behave near the crosswalks.

**Table 4.** Comparison table for the existing pedestrian safety system with our approach

| | Data sources (preprocessing automation for video) | Target situation | Characteristics of target locations | Analysis objectives | Predictive risk measure | Multi-dimensional analysis |
|---|---|---|---|---|---|---|
| Ref [55] | Pedestrian accident history (N/A) | Vehicle-pedestrian crash | All roads where accident occurred in North Carolina | Exploring the influence of factors affecting pedestrian injury severity in time period | N/A | x |
| Ref [56] | Expert views of police officers and driving public, and official road accident records (N/A) | Road accident | National road accident data in United Kingdom | Investigating the main causes of road accidents | N/A | x |
| Ref [60] | Video-based trajectory data (√) | Secondary vehicle-pedestrian interactions* | 10 unsignalized intersections in Canada | Studying the safety issue of secondary vehicle-pedestrian interactions at unsignalized intersections | x | x |
| Ref [58] | Vision-based microscopic traffic flow data(√) | Vehicle-vehicle interaction | 20 roundabouts | Analyzing behaviors and road safety proactively | x | x |
| Ref [59] | Video camera, not CCTV (x) | Vehicle-pedestrian interaction | 4 unsignalized intersections in India | Evaluating vehicle-pedestrian interactions in mixed traffic condition | x | x |
| Ref [39] | CCTV camera (video) and road characters (√) | Vehicle-pedestrian interaction | 9 different types of roads (unsignalized, signalized, school zone, etc.) in Osan City in the Republic of Korea | Analyzing vehicle-pedestrian interactions for preventing PPR** situations | x | √ |
| Proposed framework | CCTV camera (video) and road | Vehicle-pedestrian interaction | 9 different types of roads (unsignalized, | Warning the field alert for upcoming | √ (Predictive collision | √ |

| characters (√) | including predictive behaviors | signalized, school zone, etc.) in Osan City in the Republic of Korea | collisions and assessing road safety by analyzing vehicle-pedestrian interactions | risk level) |
|---|---|---|---|---|

**Note.**: √: Yes, x: No, N/A: Not applicable.
* Secondary vehicle-pedestrian interaction: interactions occurring at existing streets where vehicles leave the intersection
** PPR: Pedestrian's potential risk

In addition, the proposed framework can support to create an objective metric and movement patterns, and compare crosswalks in different conditions in multiple perspectives. However, in our experiment, we mainly focus on analyzing unsignalized crosswalks and conduct only a simple analysis for signalized crosswalks. In fact, it is difficult to handle the vehicle and pedestrian interactions without a signal phase. Thus, with the signal information, more locations could be covered, and a variety of behaviors such as signal violations can also be captured for analyses. Meanwhile, we set the PCR level by time seconds; "normal," "relatively safe," "warning," and "danger" as the overlapped time of vehicles and pedestrian's areas within 1, 2, 3, and 4 sec, respectively. These time frames can be adjusted in the PCR estimation step by decision-makers depending on road conditions such as spots where accidents occur frequently and severe weather. Our experiments handle only two scenarios, but the proposed framework can help them by adjusting levels of abstraction or adding dimensions depending on their interests or purposes.

It should be noted that the proposed framework is motivated by a lack of a comprehensive behavior analytic system for supporting administrators in making more effective decisions. Since there are more vehicle-pedestrian interactions than actual collisions, our approach can give a better and dense perspective on road safety. It recognizes patterns of potential risk situations and prevents accidents proactively. Meanwhile, as part of the proposed framework, the PCR estimation model can also serve as a field warning function to inform travelers of the upcoming collision risks. Drivers can obtain warning messages directly through the onboard units (OBUs) mounted on their vehicles, and pedestrians can also indirectly receive dynamic alerts for contextual risks using infrastructure equipment such as beam projectors. In the future, with the development of unicast communication technologies, we believe that it could be possible to send a warning message to pedestrians' mobile devices directly. The multi-dimensional analysis part could become part of a decision support system allowing administrators to study existing areas. The proposed framework, with the data accumulated multiple areas during long-term periods, can recommend alternative road designs to alleviate risky situations and test the impacts of physical or policy changes. Furthermore, with the additional data collection in urban scale, the proposed framework can provide the severity information in various administrative districts levels; crosswalk level to city level, as illustrated in **Figure 19**, and could be built in urban surveillance center.

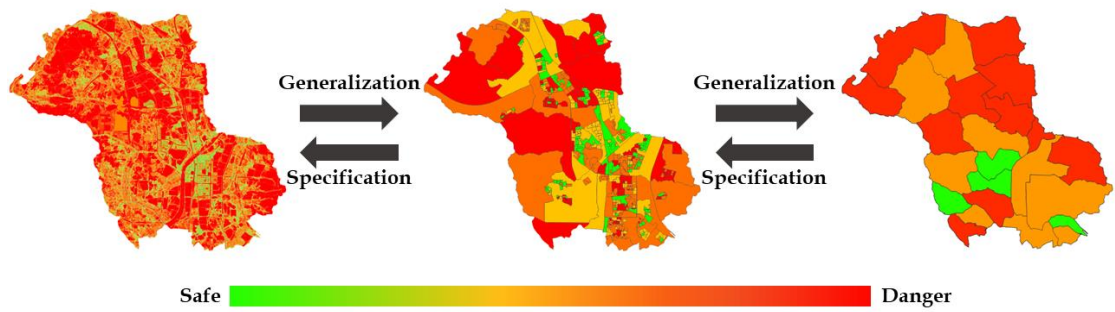

**Figure 19.** Example of severity information by administrative districts levels

## 4. Conclusions

This study proposes a newly designed analytic framework for crosswalk safety assessment using vehicle-pedestrian interactions extracted from video footage. In the preprocessing, we handle the raw video stream automatically, such as detecting and tracking objects and extracting behaviors. The core of current methodologies is to construct a data cube model combined with the PCR estimation model, called *SafetyCube2*, with a large amount of data accumulated in a data warehouse over an extended period in multiple locations. With the data cube model consisting of four hierarchical dimensions, we perform multi-dimensional analyses by using OLAP operations in various levels of abstractions. The proposed analytic framework elicits useful information that reveals the behaviors by the severity of risk levels and road environment. In addition, this framework can support decision-makers to identify risky situations, behaviors, and locations where these situations frequently occur and give clues on why such results appeared. We believe that these results work as motivations to establish the appropriate prevention and control strategies efficiently. Finally, we confirm the feasibility and applicability of the proposed framework as a decision support system for pedestrian safety by applying it to crosswalks in Osan City, Republic of Korea. Furthermore, reinforcement of the proposed analytic framework is required and it is -a part of our ongoing work.